\documentclass[10pt,twocolumn,letterpaper]{article}

\usepackage[pagenumbers]{wacv} 

\usepackage{graphicx}
\usepackage{amsmath}
\usepackage{amssymb}
\usepackage{booktabs}
\usepackage{color, colortbl}
\definecolor{Gray}{rgb}{0.9, 0.9, 0.9}
\usepackage{arydshln}

\usepackage[pagebackref,breaklinks,colorlinks]{hyperref}

\usepackage[capitalize]{cleveref}
\crefname{section}{Sec.}{Secs.}
\Crefname{section}{Section}{Sections}
\Crefname{table}{Table}{Tables}
\crefname{table}{Tab.}{Tabs.}

\def\wacvPaperID{193} 
\def\confName{WACV}
\def\confYear{2025}

\begin{document}

\title{PrevPredMap: Exploring Temporal Modeling with Previous Predictions for Online Vectorized HD Map Construction}

\author{Nan Peng\\
Ruqi Mobility\\
{\tt\small pengnan@ruqimobility.com}
\and
Xun Zhou\\
Ruqi Mobility\\
{\tt\small zhouxun@ruqimobility.com}
\and
Mingming Wang\\
GAC R\&D Center\\
{\tt\small wangmingming@gacrnd.com}
\and
Xiaojun Yang\\
Ruqi Mobility\\
{\tt\small yangxiaojun@ruqimobility.com}
\and
Songming Chen\\
GAC R\&D Center\\
{\tt\small chensongming@gacrnd.com}
\and
Guisong Chen\\
Ruqi Mobility\\
{\tt\small chenguisong@ruqimobility.com}
}
\maketitle

\begin{abstract}

   Temporal information is crucial for detecting occluded instances. Existing temporal representations have progressed from BEV or PV features to more compact query features. Compared to these aforementioned features, predictions offer the highest level of abstraction, providing explicit information. In the context of online vectorized HD map construction, this unique characteristic of predictions is potentially advantageous for long-term temporal modeling and the integration of map priors. This paper introduces PrevPredMap, a pioneering temporal modeling framework that leverages previous predictions for constructing online vectorized HD maps. We have meticulously crafted two essential modules for PrevPredMap: the previous-predictions-based query generator and the dynamic-position-query decoder. Specifically, the previous-predictions-based query generator is designed to separately encode different types of information from previous predictions, which are then effectively utilized by the dynamic-position-query decoder to generate current predictions. Furthermore, we have developed a dual-mode strategy to ensure PrevPredMap's robust performance across both single-frame and temporal modes. Extensive experiments demonstrate that PrevPredMap achieves state-of-the-art performance on the nuScenes and Argoverse2 datasets. Code will be available at \url{https://github.com/pnnnnnnn/PrevPredMap}.
   
\end{abstract}

\section{Introduction}
\label{sec:intro}


High-definition (HD) maps are essential for the autonomous driving system, providing vital information for the vehicle's localization, prediction, and planning modules. However, creating and maintaining HD maps is a labor-intensive and time-consuming endeavor. This is largely due to the necessity of achieving centimeter-level precision for various map elements and the ongoing requirement to ensure their currency. The substantial financial investment required for this process poses a significant barrier to the broader adoption and regular refresh of HD maps.


Online HD map construction is emerging as a promising alternative, offering a cost-effective solution to traditional SLAM-based methods. Utilizing onboard sensors, map elements are perceived in real-time through map learning. Early works \cite{li2022hdmapnet,pan2020cross,chen2022efficient,li2022bevformer} considered map learning as a semantic segmentation task, which required complex post-processing and often struggled with efficiency. To overcome these limitations, recent researches \cite{liu2023vectormapnet,liao2023maptr} have proposed directly predicting vectorized map elements, demonstrating a significant improvement in performance.

\begin{figure*}[!htb]
  \centering
  \includegraphics[height=6cm]{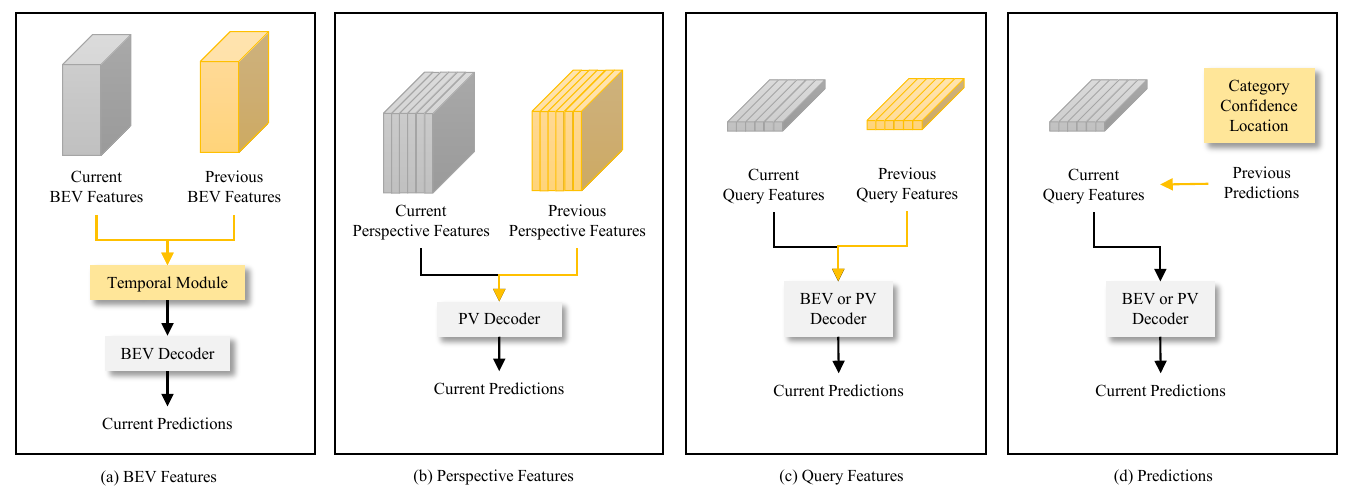}
  \caption{The simplified pipeline corresponds to various temporal representations, categorized as follows: (a) BEV features, (b) perspective features, (c) query features, and (d) predictions. Items highlighted in yellow represent the temporal modules. }
  \label{fig:motivation}
\end{figure*}


Temporal modeling is crucial for detecting occluded map elements, thereby further enhancing the quality of constructed HD maps. In the realm of Bird's-Eye-View (BEV) perception, existing temporal representations include BEV features \cite{li2022bevformer, huang2022bevdet4d, li2023bevdepth, park2022time}, perspective features \cite{liu2023petrv2, lin2022sparse4d}, and query features \cite{lin2023sparse4d, wang2023exploring, lin2023sparse4dv3}. These representations are all intermediate outcomes, as shown in \cref{fig:motivation} (a)-(c). Despite the significantly smaller information capacity of query features, the effectiveness and efficiency of temporal modeling that utilizes previous query features have been well-established \cite{lin2023sparse4d, wang2023exploring, lin2023sparse4dv3}. This progress prompts us to delve deeper: {\bf Is it still effective to rely solely on previous predictions?} Furthermore, there are at least two potential advantages to temporal modeling with previous predictions. First, predictions provide explicit information. For long-term temporal modeling, all preceding predictions can be integrated into one concise prior through advanced post-processing techniques such as filtering, tracking, merging, and curve fitting. Second, predictions are vectorized representations, facilitating the seamless integration of the framework to include map priors like standard definition (SD) maps, outdated HD maps, and crowd-sourced maps.

In this paper, we introduce our exploration of temporal modeling using previous predictions for the construction of online vectorized HD maps. This approach presents a dual challenge. On one hand, previous predictions encapsulate compact high-level information, as opposed to the rich low-level features. It is imperative that the temporal model is meticulously crafted to ensure that previous predictions are not only accurately encoded but also effectively leveraged for current predictions. On the other hand, the temporal model must excel in both single-frame and temporal modes, guaranteeing that its predictions are reliably propagated from the very first frame.

Incorporating the aforementioned considerations, we introduce PrevPredMap, a novel temporal modeling framework designed to harness previous predictions for constructing online vectorized HD maps. Two key modules of PrevPredMap are meticulously crafted: the previous-predictions-based query generator and the dynamic-position-query decoder. The previous-predictions-based query generator employs a separate encoding strategy for previous predictions, which typically encompass category, confidence, and location information. Specifically, the category and confidence information from previous predictions are encoded into the content queries, while the location information is utilized as the initial location predictions and further transformed into the initial position queries. Subsequently, the dynamic-position-query decoder introduces a dynamic update mechanism for the position queries. As depicted in \cref{fig:overview}(c), the position queries are dynamically updated based on the location predictions from the preceding decoder layer. Consequently, previous predictions are effectively encoded and utilized to produce current predictions. Moreover, we have devised a dual-mode strategy. During training, the previous-predictions-based query generator alternates randomly between single-frame and temporal modes, as shown in the upper section of \cref{fig:overview}(b). In single-frame mode, previous predictions are deliberately omitted. As a result, PrevPredMap excels in both single-frame and temporal modes during inference.

In summary, the contributions of this work include: 
\begin{itemize}
\item We introduce a novel temporal modeling framework, named PrevPredMap, which pioneers the use of previous predictions for the construction of online vectorized HD maps. 
\item Two essential modules for PrevPredMap are meticulously crafted: the previous-predictions-based query generator and the dynamic-position-query decoder. These modules not only enable PrevPredMap to effectively encode and utilize previous predictions but also ensure robust performance across both single-frame and temporal modes.
\item We present an enhanced single-frame baseline that incurs minimal computational overhead, which is fundamental to the PrevPredMap framework.
\item PrevPredMap achieves new state-of-the-art results on existing benchmarks of online vectorized HD map construction, validating the effectiveness of this innovative temporal modeling approach.
\end{itemize}

\section{Related Work}

\subsection{Online Vectorized HD Map Construction}

Online vectorized HD map construction was initially approached as a semantic segmentation task \cite{li2022hdmapnet,pan2020cross,chen2022efficient,li2022bevformer}. To build vectorized HD map, HDMapNet \cite{li2022hdmapnet} first generates BEV semantic segmentation maps and then groups these pixel-wise results to vectorized instances through heuristic post-processing. VectorMapNet \cite{liu2023vectormapnet} introduced the first end-to-end framework, utilizing an auto-regressive transformer to sequentially retrieve vectorized map instances. MapTR \cite{liao2023maptr} further advanced this end-to-end paradigm to a one-stage, parallel framework, significantly enhancing efficiency through the introduction of a unified permutation-equivalent representation and a hierarchical query embedding scheme. Concurrent and follow-up works \cite{liao2023maptrv2,hu2024admap,xu2023insightmapper,yu2023scalablemap,ding2023pivotnet,li2023lanesegnet,qiao2023end,zhang2023online,zhou2024himap,liu2024mgmap,liu2024leveraging,chen2024maptracker,gu2024accelerating} have proposed various insightful methods for further enhancement. These include the development of concise map representations \cite{ding2023pivotnet,zhang2023online,qiao2023end,yu2023scalablemap}, effective attention mechanisms \cite{liao2023maptrv2,ding2023pivotnet,hu2024admap,xu2023insightmapper}, and auxiliary supervisions \cite{liao2023maptrv2,zhang2023online,hu2024admap}. Additionally, other studies have explored the construction of vectorized HD maps using external information, such as auxiliary maps \cite{sun2023mind,luo2023augmenting,gao2023complementing,yao2023building} and temporal information \cite{yuan2024streammapnet,wang2024stream,xiong2023neural,zhu2023nemo}.

\subsection{Temporal Modeling of BEV Perception}

Temporal information for perception naturally emerges at runtime without incurring any acquisition costs. Recent advancements in BEV perception have explored the use of temporal information to enhance perception outcomes. BEVDet4D \cite{huang2022bevdet4d} and BEVFormer \cite{li2022bevformer} were the pioneers in leveraging previous BEV features for BEV perception. Subsequent works \cite{park2022time,yang2023bevformer} have expanded this temporal fusion approach from the BEV features of a single previous frame to the stacked BEV features of multiple previous frames, facilitating long-term temporal modeling. In addition, sparse-based detectors that do not rely on dense BEV features have also investigated the use of previous perspective features through deformable cross-attention mechanisms \cite{liu2023petrv2,lin2022sparse4d}. Recently, VideoBEV \cite{han2023exploring} introduced a method for propagating previous BEV features in a streaming manner, which significantly reduces storage and computation costs associated with long-term temporal modeling, as opposed to stacking multiple previous BEV features in a single forward pass. Similarly, in a streaming approach, StreamPETR \cite{wang2023exploring} and Sparse4D v2 \cite{lin2023sparse4d} have proposed utilizing previous query features, further minimizing the resource requirements for temporal information processing.


In the domain of online vectorized HD map construction, StreamMapNet \cite{yuan2024streammapnet} combines previous BEV features and query features in a streaming fashion for enhanced temporal modeling. Building upon this foundation, SQD-MapNet \cite{wang2024stream} introduces a stream query denoising strategy to learn the temporal consistency of map elements, thereby further improving the performance. Furthermore, NMP \cite{xiong2023neural} and NeMO \cite{zhu2023nemo} propose region-centric methods that harness temporal information, exhibiting significant potential for practical applications.

\begin{figure*}[!htb]
  \centering
  \includegraphics[height=13cm]{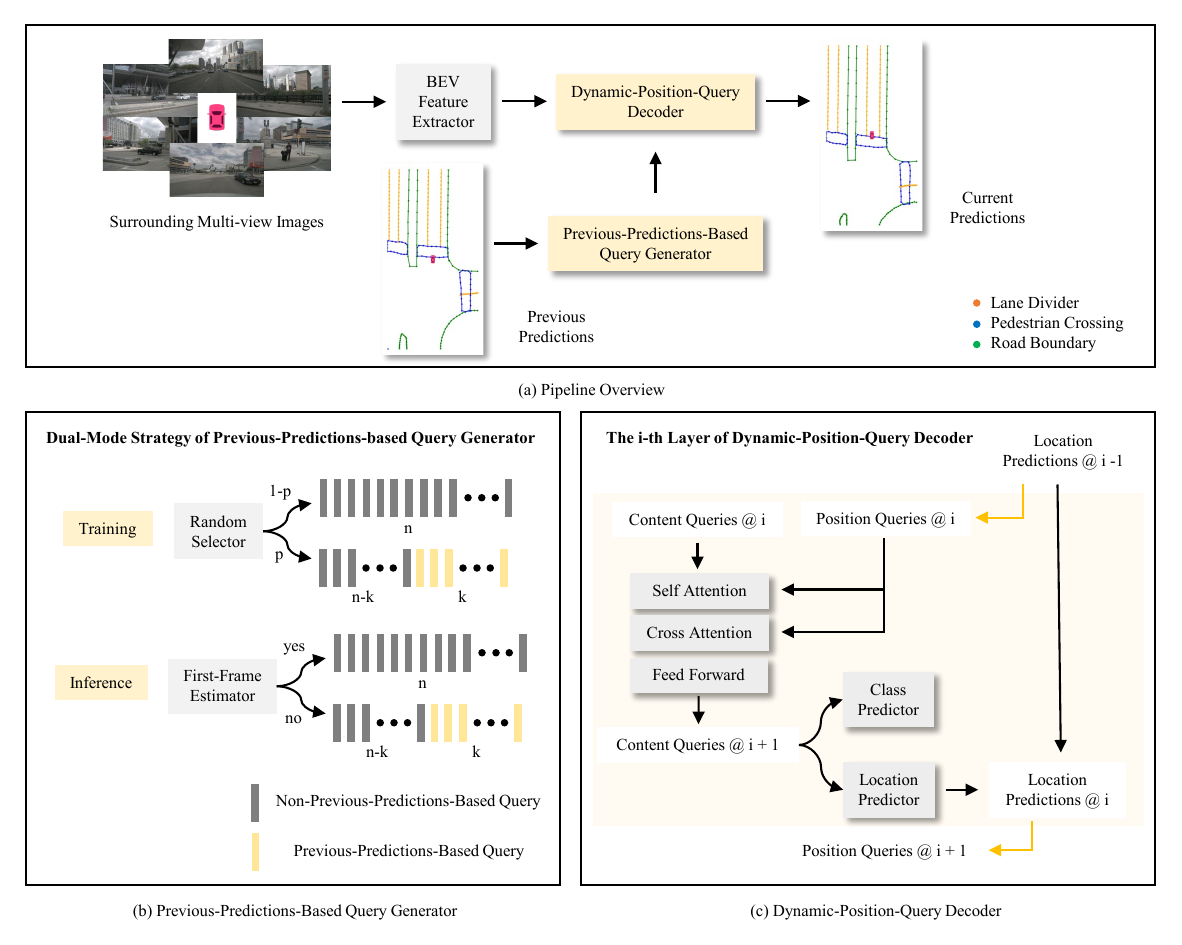}
  \caption{(a) Overall architecture of the proposed PrevPredMap, consisting of three primary modules. The BEV feature extractor is a standard part to obtain BEV features from multi-view images. The previous-predictions-based query generator and the dynamic-position-query decoder are meticulously designed to effectively encode and utilize previous predictions for producing current predictions. (b) The dual-mode strategy of the previous-predictions-based query generator. (c) The dynamic update mechanism of the dynamic-position-query decoder. Yellow arrows indicate the generation of dynamic position queries based on location predictions of the preceding decoder layer. }
  \label{fig:overview}
\end{figure*}

\section{Method}

\subsection{Overall Architecture}
PrevPredMap comprises three primary components, as illustrated in \cref{fig:overview} (a): the BEV feature extractor, the previous-predictions-based query generator, and the dynamic-position-query decoder. The BEV feature extractor is a 2D backbone followed by a parameterized PV-to-BEV transformation network, to extract BEV features from surrounding multi-view images. Subsequently, the previous-predictions-based query generator introduces to separately encode different types of information from previous predictions, which are then utilized by the dynamic-position-query decoder to produce current predictions.

\subsection{Previous-Predictions-Based Query Generator}
The previous-predictions-based query generator employs a dual-mode strategy to ensure PrevPredMap's robust performance in both temporal and single-frame modes. As illustrated in the upper section of \cref{fig:overview}(b), during training, the previous-predictions-based query generator randomly alternates between single-frame and temporal modes with probabilities $p$ and $1-p$, respectively. In single-frame mode, the queries are generated without relying on previous predictions, designated as the non-previous-predictions-based queries. In contrast, in temporal mode, the queries incorporate both the non-previous-predictions-based and previous-predictions-based queries. During inference, the previous-predictions-based query generator selects between single-frame and temporal modes based on whether the input is the first frame, as shown in the lower section of \cref{fig:overview}(b). 

In addition to the dual-mode strategy, we propose separately encoding different types of information from previous predictions. Specifically, the category and confidence information from previous predictions are encoded into the content queries, whereas the location information serves as the initial location predictions and is subsequently encoded into the initial position queries. Further details are elaborated in the two following subsections. 

\noindent {\bf Non-Previous-Predictions-Based Query.} The queries for the decoder typically include two types: content queries and position queries. The content queries are designed to extract features and generate predictions in every decoder layer, while the position queries provide the location information to the attention modules within the decoder. In line with MapTR \cite{liao2023maptr}, we employ a hierarchical query embedding scheme to encode each map element. Specifically, the hierarchical content and position queries for the j-th point of the i-th map element are formulated as: 
\begin{equation}
  q_{ij}^{con} = q_{i}^{con-ins} + q_{j}^{con-pt}, \\
  \label{eq:maptr_con}
\end{equation}
\begin{equation}
  q_{ij}^{pos} = q_{i}^{pos-ins} + q_{j}^{pos-pt}, \\
  \label{eq:maptr_pos}
\end{equation}
where $q_{i}^{con-ins}$ and $q_{i}^{pos-ins}$ are the instance-level content and position queries for the i-th map element, $q_{j}^{con-pt}$ and $q_{j}^{pos-pt}$ are the point-level content and position queries for the j-th point. Following MapTR \cite{liao2023maptr}, $q_{i}^{con-ins}$, $q_{i}^{pos-ins}$, $q_{j}^{con-pt}$, and $q_{j}^{pos-pt}$ are all learnable parameters and randomly initialized. In PrevPredMap, however, we modify the implementation of $q_{j}^{pos-pt}$ to ensure consistency between queries that are not based on previous predictions and those that are. Specifically, for the non-previous-predictions-based queries, the hierarchical content and position queries for the j-th point of the i-th map element are formulated as: 
\begin{equation}
  q_{ij}^{con} = q_{i}^{con-ins} + q_{j}^{con-pt}, \\
  \label{eq:ppmap_non_con}
\end{equation}
\begin{equation}
  q_{ij}^{pos} = q_{i}^{pos-ins} + PE(q_{j}^{loc-pt}), \\
  \label{eq:ppmap_non_pos}
\end{equation}
where $q_{j}^{loc-pt}$ represents the non-previous-predictions-based location for the j-th point, which is also learnable parameters and randomly initialized. $PE()$ denotes a position encoder, which utilizes a sinusoidal encoding function followed by a linear projection.

\noindent {\bf Previous-Predictions-Based Query.} The top k previous predictions, rationed based on confidence, are leveraged to generate previous-predictions-based queries after ego transformation. These previous predictions encompass category, confidence, and location information. By employing a separate encoding strategy, we encode the category and confidence information into the content queries, and the location information into the position queries. Specifically, for the previous-predictions-based queries, the hierarchical content and position queries for the j-th point of the i-th previous map element are formulated as:
\begin{equation}
  q_{ij}^{con} = Linear(q_{i}^{cate}) + Linear(q_{i}^{conf}) + q_{j}^{con-pt}, \\
  \label{eq:ppmap_con}
\end{equation}
\begin{equation}
  q_{ij}^{pos} = PE(q_{ij}^{loc}),
  \label{eq:ppmap_pos}
\end{equation}
where $q_{i}^{cate}$ and $q_{i}^{conf}$ are the predicted category and confidence, respectively, for the i-th previous map element. $Linear()$ denotes a linear projection function, and $q_{ij}^{loc}$ is the predicted location for the j-th point of the i-th previous map element.

\subsection{Dynamic-Position-Query Decoder}
Benefiting from the previous-predictions-based query generator, the content and position queries are encoded with previous predictions and subsequently fed into the decoder. 
\noindent {\bf Static Position Query.} In existing methods \cite{liao2023maptr,liao2023maptrv2}, the position queries remain unchanged throughout all decoder layers. Consequently, the location information supplied to the attention modules becomes outdated beyond the first decoder layer.

\noindent {\bf Dynamic Position Query.} To address the issue of outdated location information, we introduce a dynamic update mechanism that optimizes the utilization of this information. As illustrated in \cref{fig:overview} (c), the position queries at the current decoder layer are dynamically updated based on the predicted locations from the preceding decoder layer. Specifically, the position query for the j-th point of the i-th map element at the n-th decoder layer is formulated as:
\begin{equation}
q_{ij}^{pos}@n = \left\{
	\begin{array}{lll}
	q_{ij}^{pos} & if & n = 0 \\
	PE(q_{ij}^{loc}@(n-1)) & if & n > 0 \\
	\end{array}
	\right.
,
\end{equation}
where $q_{ij}^{loc}@(n-1)$ represents the predicted location for the j-th point of the i-th map element at the (n-1)-th decoder layer. Consequently, the dynamic-position-query decoder is implemented, enabling the position queries to convey up-to-date location information effectively. 

\subsection{An Enhanced Single-Frame Baseline}
Our single-frame baseline starts from MapTRv2 \cite{liao2023maptrv2}, which is open-sourced and competitive on existing benchmarks. As demonstrated by existing works \cite{liao2023maptr,li2024mapnext}, a significant improvement can be achieved by increasing the number of instance queries with only a minimal consumption of computational resources. However, MapTRv2 \cite{liao2023maptrv2} incorporates an auxiliary one-to-many matching branch to expedite convergence, which, while beneficial, results in substantial memory usage during training. Moreover, as the number of instance queries increases, the memory occupation in self-attention modules grows quadratically.

Inspired by Group DETR \cite{chen2023group}, we introduce a group-wise one-to-many branch as an innovative alternative. This approach enables the self-attention modules to operate in parallel on a group-wise basis, substantially lowering memory requirements. Moreover, the auxiliary group-wise one-to-many branch is more closely aligned with the main one-to-one branch, enhancing inference performance. Ultimately, this group-wise method, when augmented with an increased number of instance queries, forms an advanced single-frame baseline that is fundamental to PrevPredMap.

\section{Experiment}
\subsection{Experimental Setup}
\noindent {\bf Datasets.} We evaluate PrevPredMap on two popular and large-scale datasets: nuScenes \cite{caesar2020nuscenes} and Argoverse2 \cite{wilson2023argoverse}. The nuScenes dataset offers 2D vectorized maps alongside 1000 scenes, with 700 designated for training and 150 for validation. Each scene encompasses 20 seconds of 2Hz RGB images captured by 6 cameras. Argoverse2, on the other hand, delivers 3D vectorized maps and consists of 1000 logs, with 700 allocated for training and 150 for validation. Each log comprises 15 seconds of 20Hz RGB images from 7 ring cameras. To align with the Argoverse2 setup used by existing HD map construction methods, we adjust the camera frame rates from 20Hz to 2Hz and 10Hz, respectively.

\noindent {\bf Evaluation Metrics.} Consistent with previous methods \cite{li2022hdmapnet,liu2023vectormapnet,liao2023maptr}, we select three static map categories for a fair evaluation: pedestrian crossings, lane dividers, and road boundaries. The perception range is set as 30m front and rear and 15m left and right of the vehicle. The common average precision (AP) based on Chamfer Distance is used as the evaluation metric under 3 threholds of \{0.5, 1.0, 1.5\}m.

\begin{table*}[!h]
  \caption{Comparison with SOTA methods on nuScenes. All backbones utilized are ResNet50. "Temp." signifies the utilization of temporal information. The * indicates that MapTRv2 has been re-implemented with the number of instance queries set to 100. The $\star$ denotes that FPS measurements were conducted on the same machine with NVIDIA RTX A6000 for fair comparison.}
  \label{tab:nuscenes}
  \centering
  \begin{tabular}{ l c c c c c c c c c }
    \toprule
    Method & Epoch & Temp. & $AP_{div}$ & $AP_{ped}$ & $AP_{bou}$ & mAP & FPS \\
    \midrule
    MapTR \textcolor[rgb]{0.7, 0.7, 0.7}{ICLR23} \cite{liao2023maptr} & 24 & no & 51.5 & 46.3 & 53.1 & 50.3 & 15.1 \\
    MapTRv2* \textcolor[rgb]{0.7, 0.7, 0.7}{arxiv23} \cite{liao2023maptrv2} & 24 & no & 62.8 & 62.0 & 65.4 & 63.4 & 16.8$^{\star}$ \\
    StreamMapNet \textcolor[rgb]{0.7, 0.7, 0.7}{WACV24} \cite{yuan2024streammapnet} & 30 & yes & 66.3 & 61.7 & 62.1 & 63.4 & 14.9$^{\star}$ \\
    SQD-MapNet \textcolor[rgb]{0.7, 0.7, 0.7}{arxiv24} \cite{wang2024stream} & 24 & yes & 66.6 & 63.6 & 64.8 & 65.0 & 14.9$^{\star}$ \\
    \rowcolor{Gray}
    PrevPredMap (Ours) & 24 & yes & {\bf 66.9} & {\bf 64.5} & {\bf 67.6} & {\bf 66.3} & 15.7$^{\star}$ \\
    \hdashline
    MGMap \textcolor[rgb]{0.7, 0.7, 0.7}{CVPR24} \cite{liu2024mgmap} & 24 & no & 65.0 & 61.8 & 67.5 & 64.8 & 11.6 \\
    HIMap \textcolor[rgb]{0.7, 0.7, 0.7}{CVPR24} \cite{zhou2024himap} & 30 & no & {\bf 68.4} & 62.6 & {\bf 69.1} & 66.7 & 11.4 \\
    MapQR \textcolor[rgb]{0.7, 0.7, 0.7}{ECCV24} \cite{liu2024leveraging} & 24 & no & 68.0 & 63.4 & 67.7 & 66.4 & 14.7$^{\star}$ \\
    \rowcolor{Gray}
    PrevPredMap++ (Ours) & 24 & yes & 68.7 & {\bf 66.0} & 68.3 & {\bf 67.6} & 13.9$^{\star}$ \\
    \midrule
    MapTR\textcolor[rgb]{0.7, 0.7, 0.7}{ICLR23} \cite{liao2023maptr} & 110 & no & 59.8 & 56.2 & 60.1 & 58.7 & 15.1 \\
    MapTRv2* \textcolor[rgb]{0.7, 0.7, 0.7}{arxiv23} \cite{liao2023maptrv2} & 110 & no & 68.8 & 68.0 & 71.0 & 69.2 & 16.8$^{\star}$ \\
    \rowcolor{Gray}
    PrevPredMap (Ours) & 110 & yes & {\bf 70.0} & {\bf 71.2} & {\bf 72.8} & {\bf 71.3} & 15.7$^{\star}$ \\
    \bottomrule
  \end{tabular}
\end{table*}

\begin{table*}[!h]
  \caption{Comparison with SOTA methods on Argoverse2. All backbones utilized are ResNet50. "Dim." refers to the dimension of the predicted coordinates for map elements. "Freq." denotes the sampling frequency applied during training. "Temp." signifies the utilization of temporal information. The * indicates that MapTRv2 has been re-implemented with the number of instance queries set to 100. }
  \label{tab:argoverse2}
  \centering
  \begin{tabular}{ l c c c c c c c c c c }
    \toprule
    Method & Dim. & Epoch & Freq. & Temp. & $AP_{div}$ & $AP_{ped}$ & $AP_{bou}$ & mAP \\
    \midrule
    MapTRv2* \textcolor[rgb]{0.7, 0.7, 0.7}{arxiv23} \cite{liao2023maptrv2} & 2d & 30 & 2Hz & no & 70.9 & 61.7 & 65.8 & 66.1 \\
    StreamMapNet \textcolor[rgb]{0.7, 0.7, 0.7}{WACV24} \cite{yuan2024streammapnet} & 2d & 30 & 2Hz & yes & 62.0 & 59.5 & 63.0 & 61.5 \\
    SQD-MapNet \textcolor[rgb]{0.7, 0.7, 0.7}{arxiv24} \cite{wang2024stream} & 2d & 30 & 2Hz & yes & 64.9 & 60.2 & 64.9 & 63.3 \\
    \rowcolor{Gray}
    PrevPredMap(Ours) & 2d & 30 & 2Hz & yes & {\bf 72.0} & {\bf 64.2} & {\bf 68.3} & {\bf 68.2} \\
    \midrule
    MapTRv2* \textcolor[rgb]{0.7, 0.7, 0.7}{arxiv23} \cite{liao2023maptrv2} & 2d & 6 & 10Hz & no & 73.3 & 65.3 & 69.1 & 69.2 \\
    HIMap \textcolor[rgb]{0.7, 0.7, 0.7}{CVPR24} \cite{zhou2024himap} & 2d & 6 & 10Hz & no & 69.5 & {\bf 69.0} & 70.3 & 69.4 \\
    ADMapv2 \textcolor[rgb]{0.7, 0.7, 0.7}{ECCV24} \cite{hu2024admap} & 2d & 6 & 10Hz & no & 72.4 & 64.5 & 68.9 & 68.7 \\
    \rowcolor{Gray}
    PrevPredMap(Ours) & 2d & 6 & 10Hz & yes & {\bf 74.8} & 65.7 & {\bf 70.9} & {\bf 70.5} \\
    \midrule
    MapTRv2* \textcolor[rgb]{0.7, 0.7, 0.7}{arxiv23} \cite{liao2023maptrv2} & 3d & 6 & 10Hz & no & {\bf 71.7} & 62.2 & 66.7 & 66.9 \\
    \rowcolor{Gray}
    PrevPredMap(Ours) & 3d & 6 & 10Hz & yes & 71.4 & {\bf 64.1} & {\bf 67.4} & {\bf 67.6} \\
    \bottomrule
  \end{tabular}
\end{table*}

\noindent {\bf Implementation Details.} We utilize ResNet50 \cite{he2016deep} as the perspective backbone and LSS-based BEVPoolv2 \cite{huang2022bevpoolv2} as the parameterized PV-to-BEV transformation network. The optimizer is AdamW with a weight decay 0.01, and the initial learning rate is set to 0.0006, employing a cosine decay schedule. The batch size is 16 and all models are trained with 4 NVIDIA A100 GPUs. We define the size of each BEV grid as 0.3 meters. The default numbers of instance queries, point queries and decoder layers are 100, 20 and 6, respectively. For the previous-predictions-based query generator, the probability $p$ is set to 0.5. The amount of the previous-predictions-based queries, denoted as $k$, is set to 10 for nuScenes and 8 for Argoverse2. For auxiliary group-wise one-to-many matching, the number of groups is 6. During training, the top $k$ predictions of all training samples are stored and updated immediately after their respective iterations in a predefined dictionary, allowing for the retrieval of previous predictions as needed.

\subsection{Comparisons with State-of-the-art Methods}

\noindent {\bf Performance on nuScenes.} As depicted in \cref{tab:nuscenes}, PrevPredMap achieves 66.3 mAP and 71.3 mAP with 24-epoch and 110-epoch training schedules, respectively. These results outperform all SOTA methods in terms of training convergence, validation accuracy and inference speed. When enhanced with GKT-h, a more powerful BEV encoder introduced in MapQR \cite{liu2024leveraging}, PrevPredMap++ attains 67.5 mAP with a 24-epoch training schedule.

\noindent {\bf Performance on Argoverse 2.} Argoverse2 offers a 3D vectorized map, which includes additional height information not present in the nuScenes dataset. Since most existing works do not utilize this height information, with the exception of MapTRv2 \cite{liao2023maptrv2}, we train PrevPredMap under both 2D and 3D configurations. For fair comparison, the sampling frequency is set to 2Hz and 10Hz, respectively. It should be noted that the amount of training data at a 2Hz sampling frequency is 20$\%$ of that at a 10Hz sampling frequency. As observed in \cref{tab:argoverse2}, PrevPredMap surpasses all SOTA methods in both 2D and 3D vectorized map construction on Argoverse2. 

\subsection{Ablation Study}

\begin{table*}[tb]
  \caption{Ablation of the proposed PrevPredMap. The experiments were performed on nuScenes, following a 24-epoch training schedule. FPS measurements are conducted on the same machine with NVIDIA RTX A6000. }
  \label{tab:component}
  \centering
  \begin{tabular}{ c c c c c c }
    \toprule
    Query & Group-Wise & Previous-Predictions-based & Dynamic-Position-Query & mAP & FPS \\
    Amount & One-to-Many & Query Qenerator & Decoder &  &  \\
    \midrule
    50 &  &  &  & 61.5 & 16.8 \\
    100 &  &  &  &  63.4 & 16.8 \\
    100 & \checkmark &  &  &  64.8 & 16.8 \\
    100 & \checkmark & \checkmark &  &  65.4 & 16.1 \\
    100 & \checkmark &  & \checkmark &  64.9 & 16.0 \\
    \rowcolor{Gray}
    100 & \checkmark & \checkmark & \checkmark &  66.3 & 15.7 \\
    \bottomrule
  \end{tabular}
\end{table*}

\noindent {\bf Components Ablation.} As demonstrated in \cref{tab:component}, a straightforward increase in the query amount from 50 to 100, complemented by an auxiliary group-wise one-to-many branch, achieves a total improvement of 3.3 mAP. Additionally, we independently assess the impact of the previous-prediction-based query generator and the dynamic-position-query decoder. The individual enhancements from these modules are modest. However, when integrated, these modules effectively encode and utilize previous predictions, leading to a final enhancement of 4.8 mAP, accompanied by only a 6.5$\%$ reduction in FPS.

\begin{table}[tb]
  \caption{Performance comparison of the single-frame and temporal modes for PrevPredMap on nuScenes, following a 24-epoch training schedule. FPS measurements are conducted on the same machine with NVIDIA RTX A6000.}
  \label{tab:circumstance}
  \centering
  \begin{tabular}{ l c c c c c }
    \toprule
    Mode & mAP & $AP_{div}$ & $AP_{ped}$ & $AP_{bou}$ & FPS \\
    \midrule
    Single-Frame & 65.7 & 66.0 & 63.9 & 67.3 & 16.1 \\
    Temporal & 66.3 & 66.9 & 64.5 & 67.6 & 15.7 \\
    \bottomrule
  \end{tabular}
\end{table}

\noindent {\bf Performance Comparison of the Single-Frame and Temporal Modes.} Benefiting from the proposed dual-mode strategy implemented in the previous-predictions-based query generator, PrevPredMap excels in both single-frame and temporal modes. As depicted in \cref{tab:circumstance}, the single-frame mode of PrevPredMap is 65.7 mAP, which is still comparable to the existing state-of-the-art methods listed in \cref{tab:nuscenes}. This characteristic is advantageous because online inference can be interrupted due to various unexpected emergencies. It is possible that the single-frame mode may occasionally be called upon during runtime. Additionally, the discrepancy between the single-frame and temporal modes is not particularly noticeable. We offer two possible explanations for this phenomenon. First, map elements are more likely to be severely occluded in situations such as when a large truck is nearby or during a traffic jam; however, these scenarios are rare in existing datasets, thereby limiting the advantage of the temporal mode. As shown in Table 3 of SQD-MapNet \cite{wang2024stream}, the mAP even decreases by 0.7 mAP when the temporal stream is added to the single-frame baseline. Second, PrevPredMap currently only utilizes predictions from the previous frame. As demonstrated in Table 5 of StreamPETR \cite{wang2023exploring}, there is a substantial increase in mAP if the number of frames considered is increased from one to seven. We reserve this avenue for future exploration, as the post-processing of predictions from multiple preceding frames is complex and would require significant effort in refining detailed settings. Finally, the dual-mode setting for temporal modeling, as introduced by PrevPredMap, is unprecedented. There are no existing papers for direct comparison.

\begin{table}[t]
  \caption{Probability of the previous-predictions-based query generator. All experiments were conducted on nuScenes, following a 24-epoch training schedule.}
  \label{tab:probability}
  \centering
  \begin{tabular}{ c c c c c }
    \toprule
    Probability & mAP & $AP_{div}$ & $AP_{ped}$ & $AP_{bou}$ \\
    \midrule
    0.4 & 66.2 & 65.8 & {\bf 65.4} & 67.3 \\
    0.5 & {\bf 66.3} & {\bf 66.9} & 64.5 & {\bf 67.6} \\
    0.6 & 65.7 & 65.8 & 63.8 & 67.5 \\
    \bottomrule
  \end{tabular}
\end{table}

\noindent {\bf Probability of the Previous-Predictions-Based Query Generator.} As depicted in the upper part of \cref{fig:overview} (b), the probabilities of $p$ and $1-p$ represent the propensity for temporal and single-frame learning, respectively. According to \cref{tab:probability}, PrevPredMap attains optimal performance at a probability of 0.5, likely owing to the equilibrium established between temporal and single-frame learning modes in its training regimen.

\begin{table}[tb]
  \caption{The number of previous-predictions-based queries. All experiments were conducted on nuScenes, following a 24-epoch training schedule.}
  \label{tab:number}
  \centering
  \begin{tabular}{ c c c c c }
    \toprule
    k & mAP & $AP_{div}$ & $AP_{ped}$ & $AP_{bou}$ \\
    \midrule
    5 & 66.1 & 65.5 & {\bf 65.0} & 67.5 \\
    8 & 66.3 & 65.8 & 64.9 & {\bf 68.2} \\
    10 & 66.3 & {\bf 66.9} & 64.5 & 67.6 \\
    12 & {\bf 66.5} & {\bf 66.9} & 64.5 & 68.0 \\
    15 & 65.7 & 65.3 & 64.8 & 66.9 \\
    \bottomrule
  \end{tabular}
\end{table}

\begin{figure*}[!htb]
  \centering
  \includegraphics[height=11cm]{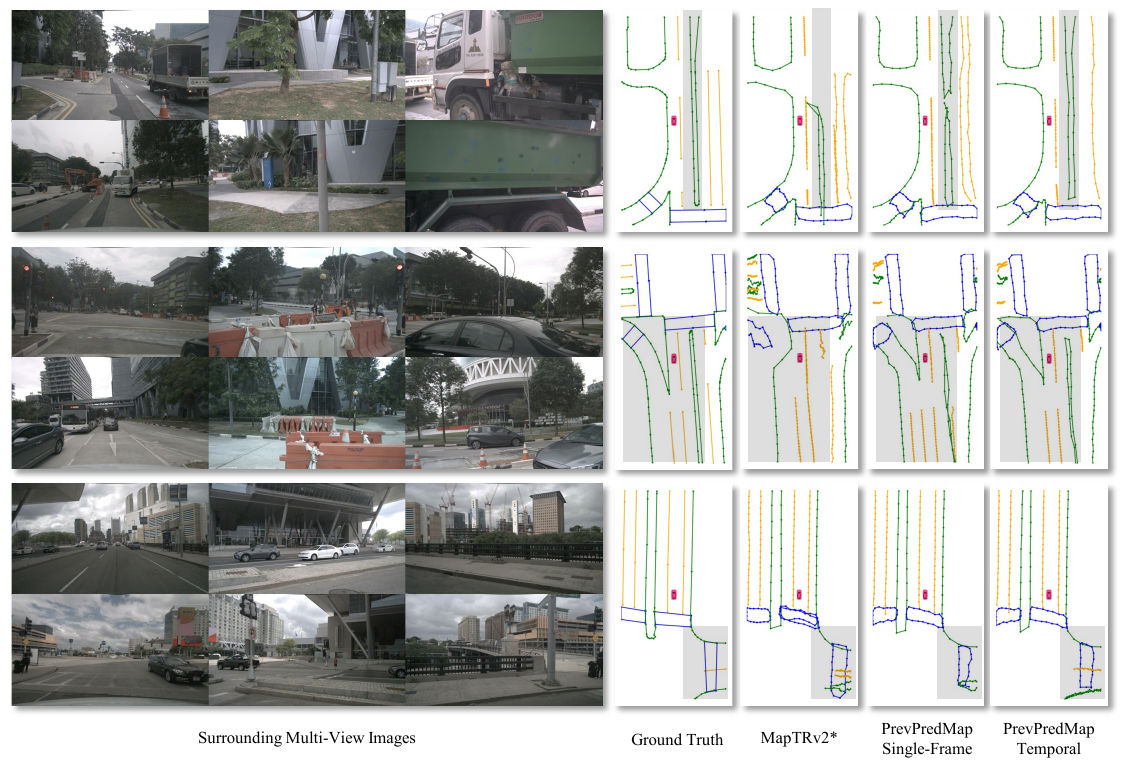}
  \caption{Comparison of PrevPredMap with single-frame SOTA methods on qualitative visualization under various occlusion scenarios. Each sub-part displays four qualitative results: Ground Truth, MapTRv2, PrevPredMap Single-Frame, and PrevPredMap Temporal. The * indicates that MapTRv2 has been re-implemented with the number of instance queries set to 100. Green, orange and blue lines denote road boundaries, lane dividers and pedestrian crossings, respectively.}
  \label{fig:qa}
\end{figure*}

\noindent {\bf The Number of Previous-Predictions-Based Queries.} As illustrated in \cref{tab:number}, PrevPredMap's performance increases with an elevated count of previous-predictions-based queries, reaching saturation at a number approximately 12. It is important to note that the average number of map elements per frame is less than 10 for nuScenes. An excessively high value of $k$ might introduce a multitude of previous predictions characterized by high uncertainty, potentially leading to adverse effects.

\noindent {\bf Qualitative Analysis.} \cref{fig:qa} displays three common scenarios in autonomous driving, where the camera's field of view is partially obstructed by a large truck, plastic water-filled barriers, and roadside railings, respectively. As shown in the highlighted gray areas of \cref{fig:qa}, single-frame models struggle to accurately perceive map elements due to the various obstacles. In contrast, PrevPredMap effectively leverages previous predictions for constructing vectorized HD maps, underscoring the effectiveness of this innovative temporal modeling approach.

\subsection{Limitations and Future Work}
Based on our current understanding, the limitations and future work of PrevPredMap are discussed in two main aspects. Firstly, PrevPredMap currently leverages predictions solely from the preceding frame. It is anticipated that the performance of the temporal mode could be significantly enhanced by adeptly post-processing and utilizing predictions from multiple preceding frames. Secondly, map priors constitute essential complementary information for online HD map construction. There is a strong expectation that incorporating map priors into the PrevPredMap framework will further enhance the reliability of the constructed map. 

\section{Conclusion}
This paper focuses on temporal modeling using previous predictions for the construction of online vectorized HD maps. We introduce two meticulously designed modules: the previous-predictions-based query generator and the dynamic-position-query decoder. With these, we propose a novel temporal modeling framework, PrevPredMap, which leverages previous predictions to enhance online vectorized HD map construction. These modules not only enable PrevPredMap to effectively encode and utilize previous predictions but also ensure robust performance across both single-frame and temporal modes. Besides, we present an enhanced single-frame baseline that incurs minimal computational overhead, which is fundamental to the PrevPredMap framework. PrevPredMap is simple yet effective, achieving new state-of-the-art results on existing benchmarks. We hope that PrevPredMap will provide new insights into the temporal modeling of BEV perception for the community. 

{\small
\bibliographystyle{ieee_fullname}
\bibliography{egbib}
}

\end{document}